\documentclass[preprint,review,12pt]{elsarticle}

\usepackage{amssymb}
\usepackage{amsmath}

\usepackage{lineno}

\journal{Knowledge-Based Systems}
\usepackage[dvipsnames]{xcolor}
\usepackage{bbm}
\usepackage{tabularx}
\usepackage{multirow}
\usepackage[linesnumbered,ruled]{algorithm2e}

\newcommand{\image}{\mathbf{X}}
\newcommand{\feature}{\mathbf{F}}

\newcommand{\dstrain}{\mathcal{D}^{src}}
\newcommand{\dttrain}{\mathcal{D}^{tar}}
\newcommand{\dtest}{\mathcal{D}^{test}}

\newcommand{\stask}{\mathcal{T}^{src}}
\newcommand{\ttask}{\mathcal{T}^{tar}}

\newcommand{\lfsl}{\mathcal{L}_{fsl}}

\newcolumntype{l}{>{\centering\arraybackslash\hsize=1.0\hsize}X}
\newcolumntype{m}{>{\hsize=1.0\hsize}X}
\newcolumntype{s}{>{\hsize=1.0\hsize}X}

\newcommand{\NB}[1]{\noindent\textbf{#1}}

\usepackage{hyperref}

\begin{document}
\begin{frontmatter}



\title{FreqGRL: Suppressing Low-Frequency Bias and Mining High-Frequency Knowledge for Cross-Domain Few-Shot Learning} 



\author[Xjtu1]{Siqi Hui} 
\author[Xjtu1]{Sanping Zhou}
\author[Xncj]{Ye Deng}
\author[Nbgc]{Wenli Huang}
\author[Xjtu1]{Yang Wu}
\author[Xjtu1]{Jinjun Wang\corref{cor1}}
\cortext[cor1]{Corresponding author}
\ead{jinjun@mail.xjtu.edu.cn}

\affiliation[Xjtu1]{organization={National Key Laboratory of Human-Machine Hybrid Augmented Intelligence},addressline={Xian Ning West Road No.28}, 
            city={Xi'an},
            postcode={710049}, 
            state={Shaanxi},
            country={China}}
\affiliation[Xncj]{organization={Engineering Research Center of Intelligent Finance, Ministry of Education, School of Computing and Artificial Intelligence, Southwestern University of Finance and Economics}, addressline={LiutaiAvenue No.555}, city={Chengdu}, postcode={611130}, state={Sichuan}, country={China}}
\affiliation[Nbgc]{organization={School of Electronic and Information Engineering, Ningbo University of Technology}, addressline={No. 201, Fenghua Road, Jiangbei District}, city={Ningbo}, postcode={315211}, state={Zhejiang}, country={China}}
\begin{abstract}
Cross-domain few-shot learning (CD-FSL) aims to recognize novel classes with only a few labeled examples under significant domain shifts. While recent approaches leverage a limited amount of labeled target-domain data to improve performance, the severe imbalance between abundant source data and scarce target data remains a critical challenge for effective representation learning. We present the first frequency-space perspective to analyze this issue and identify two key challenges: (1) models are easily biased toward source-specific knowledge encoded in the low-frequency components of source data, and (2) the sparsity of target data hinders the learning of high-frequency, domain-generalizable features. To address these challenges, we propose \textbf{FreqGRL}, a novel CD-FSL framework that mitigates the impact of data imbalance in the frequency space. Specifically, we introduce a Low-Frequency Replacement (LFR) module that substitutes the low-frequency components of source tasks with those from the target domain to create new source tasks that better align with target characteristics, thus reducing source-specific biases and promoting generalizable representation learning. We further design a High-Frequency Enhancement (HFE) module that filters out low-frequency components and performs learning directly on high-frequency features in the frequency space to improve cross-domain generalization. Additionally, a Global Frequency Filter (GFF) is incorporated to suppress noisy or irrelevant frequencies and emphasize informative ones, mitigating overfitting risks under limited target supervision. Extensive experiments on five standard CD-FSL benchmarks demonstrate that our frequency-guided framework achieves state-of-the-art performance.
\end{abstract}




\end{frontmatter}
\section{Introduction}
Few-shot learning (FSL) enables models to learn novel concepts from only a limited number of labeled examples~\cite{sung2018learning,gao2022label,hospedales2021meta}. However, most existing FSL approaches assume that training and testing data originate from the same domain—a restriction that is often impractical in real-world scenarios. To address this limitation, Cross-domain few-shot learning (CD-FSL) has recently emerged, extending the foundational principles of FSL to facilitate knowledge transfer across heterogeneous domains. CD-FSL has demonstrated significant advancements in diverse applications.

Traditionally, CD-FSL methods rely exclusively on labeled samples from the source domain for model training, without access to target-domain data~\cite{wang2021cross,fu2022wave}. However, since the source data alone cannot adequately capture the diverse distributional characteristics of potential target domains, these methods often suffer from performance degradation in practical applications. To alleviate this limitation, recent studies have incorporated unlabeled target-domain data to learn more domain-adaptive representations~\cite{fu2021meta,oh2022refine}. Nonetheless, acquiring sufficient target-domain data remains challenging when events occur infrequently. Notably, several recent approaches~\cite{hui2024gradient,hui2024mcm,li2025cdcnet} have demonstrated competitive performance by leveraging only a few labeled samples from the target domain. Since collecting and annotating a small amount of target data is typically feasible in practice, we adopt this more realistic and practically viable setting.


We highlight two critical challenges that persist in scenarios with limited labeled target data: \textbf{First}, substantial domain shifts between the source and target domains often cause models to overly rely on abundant source data, acquiring \textbf{source-specific} knowledge that impairs generalization to the target domain. \textbf{Second}, the scarcity of labeled target samples restricts the model from learning expressive and \textbf{domain-generalizable} representations. Moreover, it increases the risk of overfitting to noisy or spurious patterns, particularly when training overemphasizes the few available samples, ultimately compromising the robustness and stability of domain adaptation. 

Addressing these challenges is crucial for advancing CD-FSL in realistic, data-scarce scenarios. Existing methods primarily tackle these issues through spatial-domain strategies such as data augmentation~\cite{fu2021meta,zhuo2022tgdm}, feature disentanglement~\cite{fu2022me,fu2021meta}, adversarial learning~\cite{fu2021meta}, and various forms of regularization techniques~\cite{hui2024mcm,hui2024gradient}. Although effective to some extent, these approaches predominantly rely on pixel-level representations, which inherently blend both domain-specific and domain-invariant features due to their direct encoding of visual appearance. This blending makes it challenging for models to explicitly separate domain-specific and domain-invariant features. In contrast, frequency-domain modeling offers a promising alternative, enabling more explicit disentanglement of these features, which remains underexplored in the current literature.


\begin{figure}[t]
    \centering
    \includegraphics[width=0.7\linewidth]{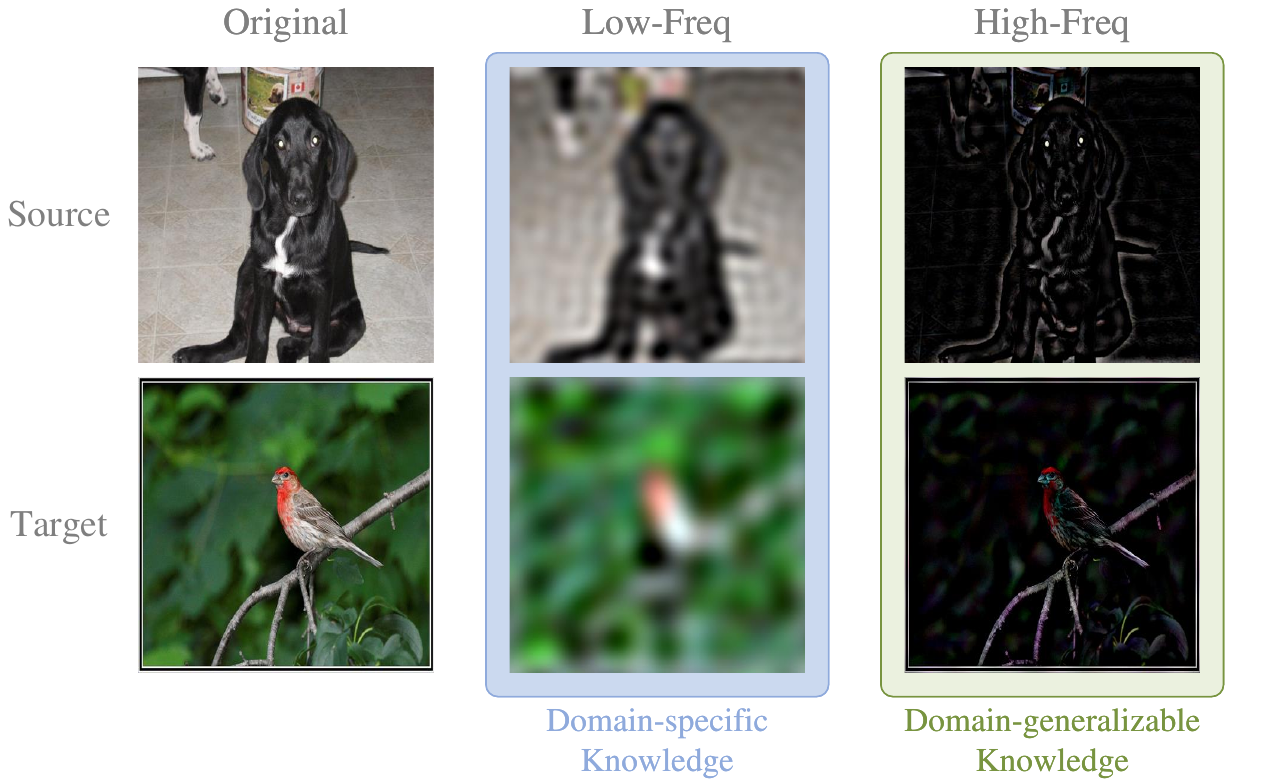}
    \vspace{-0.5cm}
    \caption{
    Illustration of image reconstruction using low-frequency and high-frequency components on both the source dataset (miniImagenet) and target dataset (CUB-2011-200). Original images are first transformed into frequency domain via the Fast Fourier Transformation (FFT), then decomposed into their respective low-frequency and high-frequency components. Subsequently, images containing only low- or high-frequency information are reconstructed by inversely mapping these individual components back into the original spatial domain. 
    }
    \label{fig:freq-recon}
\vspace{-0.5cm} \end{figure}
\begin{figure}[t]
    \centering
    \includegraphics[width=0.7\linewidth]{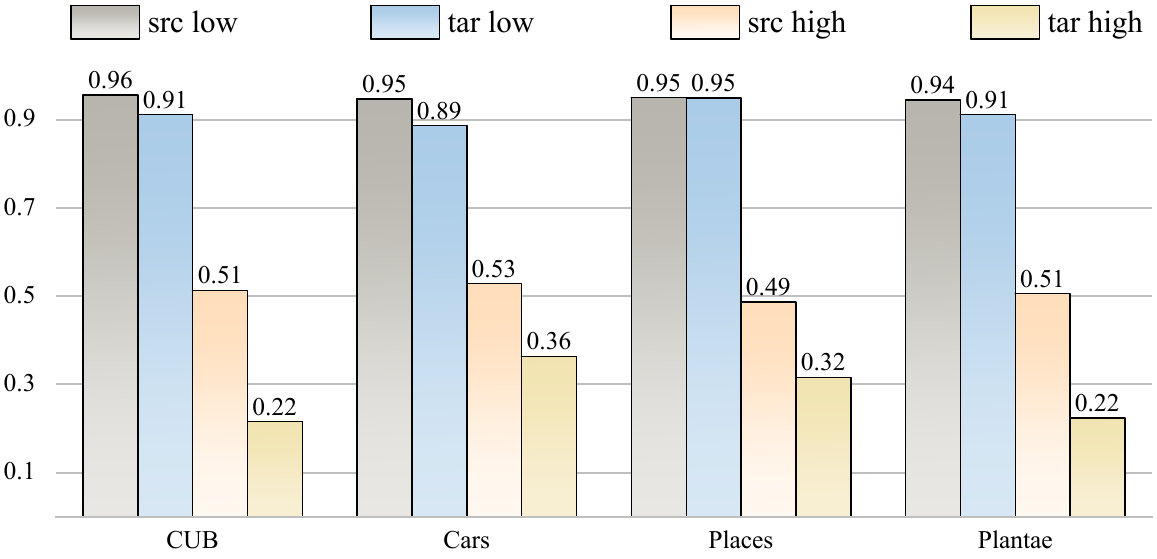}
    \vspace{-0.5cm}
    \caption{Accuracy ratios of full-spectrum-trained models evaluated on tasks reconstructed from specific frequency components. Each bar represents the ratio of the model's accuracy on frequency-specific tasks (tasks reconstructed using only low-frequency or high-frequency components) relative to its accuracy on the original tasks. Results are reported for four target datasets using a ResNet-10 backbone, with models trained on a combination of source-domain data and limited labeled target-domain data.}
    \label{fig:high-low-acc}
\vspace{-0.5cm} \end{figure}

Different frequency components encode distinct characteristics, some of which capture domain-generalizable features that can enhance the generalization~\cite{chen2021few,qin2024fdgnet}. Typically, frequency components can be categorized into low-frequency and high-frequency signals~\cite{chen2023learnable,guo2023aloft}. Low-frequency signals predominantly encode domain-specific attributes, such as textures, styles, and background elements. In contrast, high-frequency signals primarily represent semantic information, such as edges and contours, which are less sensitive to domain variations and thus more beneficial for domain generalization~\cite{ji2025frequency,liu2024spectral,ying2025high}. 

To illustrate this clearly, we present frequency decomposition and image reconstruction experiments using low-frequency or high-frequency components in Figure~\ref{fig:freq-recon}. For deeper insights into the impact of frequency components from source and target domains, we further evaluate model performance on tasks reconstructed solely from either low- or high-frequency signals. As demonstrated in Figure~\ref{fig:high-low-acc}, we make two key observations: 1) On the \textbf{source domain}, the model achieves notably higher accuracy on tasks reconstructed from low-frequency components compared to those reconstructed from high-frequency components, with accuracy of 0.96 (CUB~\cite{branson2010visual}), 0.95 (Cars~\cite{krause20133d}), 0.95 (Places~\cite{zhou2017places}), and 0.94 (Plantae~\cite{van2018inaturalist}). In contrast, the accuracy for high-frequency tasks are significantly lower, at only 0.51, 0.53, 0.49, and 0.51, respectively. This substantial gap indicates that the model predominantly relies on low-frequency components in the source domain, which typically encode domain-specific patterns. Such reliance poses the risk of embedding excessive domain-specific knowledge from the source data, potentially hindering generalization to target domains; 2) Regarding the \textbf{target domain}, the accuracy on high-frequency tasks drops significantly compared to the source domain, falling to 0.22 (CUB), 0.36 (Cars), 0.32 (Places), and 0.22 (Plantae), with an average of only 0.28-markedly lower than the 0.51 average observed in the source-domain. This notable discrepancy indicates that the model struggles to acquire high-frequency, domain-generalizable knowledge from the limited target-domain data. These results highlight a clear asymmetry in how frequency components are leveraged across domains, underscoring the necessity for frequency-aware strategies capable of suppressing source-specific low-frequency biases while effectively capturing generalizable high-frequency representations.

Building on these insights, we propose a novel framework, named \textbf{FreqGRL}uency-Guided \textbf{G}eralization \textbf{R}epresentation \textbf{L}earning FreqGRL, which aims to suppress the learning of source-specific knowledge while enhancing the extraction of generalizable features from a frequency domain perspective. As a core component, we introduce a Fast Fourier Transform (FFT)-based data augmentation module, termed \textbf{Low-Frequency Replacement (LFR)}, designed to reduce the influence of source-specific information. Specifically, LFR replaces the low-frequency components of each class in the source domain with those from a corresponding class in the target domain. By doing so, it alleviates domain shifts and encourages the model to focus on learning domain-invariant features. Additionally, we propose the \textbf{High-Frequency Enhancement (HFE)} module to strengthen the model’s focus on high-frequency, domain-generalizable information. HFE applies a channel-wise FFT to intermediate feature maps to obtain their frequency representations, then selectively masks out low-frequency components while preserving only the high-frequency ones. Unlike prior approaches that remap the frequency signals back into the spatial domain for further processing, our HFE performs convolutions directly in the frequency domain. This design is particularly, as manipulating specific frequency components in the frequency space produces global effects on the corresponding spatial feature before the FFT, thereby facilitating more efficient learning of high-level semantic representations. Finally, we introduce the \textbf{Global Frequency Filter (GFF)}, which helps the network in enhancing informative components while suppressing noise present in the limited target-domain data. GFF applies a channel-wise FFT to each feature map and performs an element-wise multiplication between the resulting frequency representation and a learnable global weighting tensor in the frequency domain. This design enables dynamic, end-to-end modulation of frequency components, thereby promoting more effective cross-domain generalization throughout the training process.

The key contributions of this paper are as follows:
\vspace{-0.4cm}
\begin{itemize}
    \item We reveal the asymmetric use of frequency components in CD-FSL and identify low-frequency bias and insufficient high-frequency learning as key obstacles to generalization.
    \vspace{-0.4cm}
    \item We propose FreqGRL, a novel frequency-guided framework that mitigates low-frequency bias via Low-Frequency Replacement (LFR) and enhances generalizable features through High-Frequency Enhancement (HFE) and Global Frequency Filtering (GFF).
    \vspace{-0.4cm}
    \item  Extensive experiments on four benchmarks demonstrate that FreqGRL achieves new state-of-the-art results in both 1-shot and 5-shot CD-FSL, outperforming existing methods by a clear margin.
\end{itemize}
\section{Related Work}
\subsection{Few-shot learning}

Inspired by human cognitive processes, FSL aims to enable models to learn novel concepts from a limited number of labeled samples. Existing methods can be broadly categorized into metric-based and meta-learning-based methods. Metric-based methods learn a discriminative embedding space where classification is performed by measuring the distances between support and query samples using predefined metric function~\cite{koch2015siamese,vinyals2016matching,snell2017prototypical}. In contrast, meta-learning-based methods aim to learn a generalizable model initialization that can be quickly adapted to new tasks with only a few optimization steps~\cite{vilalta2002perspective,triantafillou2019meta}. However, both categories typically assume that the training and testing data are drawn from the same domain, which limits their performance in scenarios involving domain shifts. 

\subsection{Cross-domain few-shot learning}
CD-FSL aims to enable few-shot learning in the presence of significant domain shifts between the source and target domains. Depending on the availability of target-domain data, CD-FSL methods can be broadly categorized into three settings: target-free, unlabeled target, and limited labeled target approaches. Target-free methods rely solely on the source-domain data for training, with no access to target-domain information. To simulate target-domain variability, these methods often employ task augmentation~\cite{wang2021cross}, feature perturbation (e.g., wave-SAN~\cite{fu2022wave}, NSAE~\cite{liang2021boosting}, FWT~\cite{tseng2020cross}, AFA~\cite{hu2022adversarial}), or adversarial training techniques~\cite{fu2023styleadv}. In some cases, fine-tuning is applied on the target support set~\cite{das2021importance,guo2020broader,li2022ranking,liang2021boosting}, though the absence of target information during training often limits generalization performance. Unlabeled target methods address this limitation by incorporating unlabeled target data into the training process, leveraging self-supervised learning to extract domain-relevant features~\cite{phoo2020self,yao2021cross,liu2023self}. However, acquiring large volumes of unlabeled data is not always feasible in practice. Limited labeled target methods have thus gained traction as a more practical alternative. In this setting, a small number of labeled samples from the target domain are available during training. Recent approaches leverage this scarce supervision through task augmentation (e.g., FDMixup~\cite{fu2022me}, TGDM~\cite{zhuo2022tgdm}, CDCNet~\cite{li2025cdcnet}) and regularization techniques based on domain disentanglement (e.g., Me-D2N~\cite{fu2022me}, GGCM~\cite{hui2024gradient}, and MCM~\cite{hui2024mcm}) to bridge the source–target gap. In this paper, we adopt the limited labeled target setting. However, unlike existing methods, we investigate and address its core challenges from a novel frequency-based perspective, aiming to suppress source-specific bias and enhance domain-generalizable representation learning.

\subsection{Frequency learning}
Frequency analysis has long served as a fundamental tool in traditional digital image processing and has recently gained increasing attention in deep learning-based computer vision tasks, including compressed sensing~\cite{wang2018packing}, visual pre-training~\cite{xie2022masked}, semantic segmentation~\cite{huang2021fsdr}, and domain adaptation~\cite{huangfrequency}. Beyond performance enhancement, frequency analysis has also been used to interpret deep neural networks, such as analyzing their inductive biases~\cite{geirhos2018imagenet} or identifying the specific frequency components to which they are most responsive during learning~\cite{dziedzic2019band,xu2020learning}. Recent studies have increasingly leveraged frequency-domain representations to improve model generalization. Some works incorporate frequency spectra as complementary signals to spatial features, thereby enhancing feature discriminability~\cite{xu2020learning,chen2021few,li2022few,li2025fspdf}. Others highlight the strength of frequency-domain processing in capturing non-local receptive fields, enabling models to extract high-level structural information~\cite{ji2025frequency}. In particular, high-frequency components have been shown to encode semantically rich patterns that are beneficial for generalization~\cite{luo2021generalizing,bai2022improving}. For example, Lin et al.~\cite{liu2024spectral} improve robustness by amplifying high-frequency components while suppressing low-frequency ones. In the few-shot setting, FicNet~\cite{zhu2024few} incorporates frequency cues to capture compact structural patterns, while Wave-SAN~\cite{fu2022wave} introduces frequency-based modifications via discrete wavelet transform to enhance feature style in target-free CD-FSL. Other methods perform frequency masking as a form of task augmentation to simulate domain variability~\cite{liu2024spectral}. In contrast to these approaches, we explore frequency modeling in the context of CD-FSL with limited labeled target data, addressing its unique challenges from a frequency-aware perspective. Specifically, our proposed LFR module performs task augmentation guided by the target task to selectively suppress source-specific information, unlike prior methods~\cite{fu2022wave,liu2024spectral} that operate solely on source tasks. Furthermore, the HFE module enhances the model’s sensitivity to high-frequency components in intermediate features, enabling more effective extraction of domain-generalizable cues from sparse target data.
\section{Method}
\vspace{-0.15cm}\subsection{Preliminary}
\NB{Problem formulation.}
Conventional CD-FSL methods aim to train models exclusively on a source dataset $\dstrain$ \ and transfer the learned knowledge to a target dataset $\dtest$. In our work, we adopt a more practical setting by introducing an auxiliary target dataset $\dttrain$, which provides a small number of labeled target-domain data during training.

In the standard CD-FSL paradigm, models are trained and evaluated on \emph{N}-way \emph{K}-shot tasks. Each task $\mathcal{T}=\{S,Q\}$ comprises of a support set $S=\{(x_i,y_i)\}_{i=1}^{N \times K}$ consisting of $N$ classes with $K$ labeled samples per class, and a query set $Q=\{(x_i,y_i)\}_{i=1}^{N \times M}$, which includes $M$ unlabled images per class drawn from the same classes as the support set. Tasks sampled from $\dstrain$, $\dttrain$, and $\dtest$ are denoted as $\mathcal{T}^{src}=\{S^{src},Q^{src}\}$, $\mathcal{T}^{tar}=\{S^{tar},Q^{tar}\}$, and $\mathcal{T}^{test}=\{S^{test},Q^{test}\}$, respectively. 

During training, model is optimized using the few-shot loss $\lfsl$ \ on both the source dataset $\dstrain$ and the target training dataset $\dttrain$. Specifically, given a source training task $\stask$ containing $N\times M$ unlabeled images, the few-shot classification loss can be defined as $\lfsl^{src}=\sum_{i=1}^{NM}\mathcal{L}_{ce}(p^{src}_i,y^{src}_i)$, where $p^{src}_i$ denotes the predicted class probability of the \emph{i}-th query sample, and $\mathcal{L}_{ce}$ is the cross-entropy loss function. The few-shot loss on the target training dataset is defined in the same manner.

\NB{Frequency transformation.}
In this section, we briefly introduce a conventional signal processing tool, the 2D Fast Fourier Transform (FFT). We apply the 2D FFT to both input images and intermediate feature maps, enabling frequency-domain analysis and learning.

Formally, given an input image $\image\in\mathbb{R}^{3\times H \times W}$, we apply the 2D FFT independently to each channel to obtain its corresponding frequency spectrum $\image_f$, computed as:

\begin{equation}
\mathbf{X}_f(x, y) = \sum_{h=0}^{H-1} \sum_{w=0}^{W-1} \mathbf{X}(h, w) e^{-j 2\pi \left( x\frac{h}{H} + y\frac{w}{W} \right)}.
\end{equation}
The frequency spectrum $\image_f$ can be transformed back to the original spatial space via the inverse FFT $iFFT(\cdot)$, defined as follows:

\begin{equation}
\mathbf{X}(h, w) = \frac{1}{H \cdot W} \sum_{x=0}^{H-1} \sum_{y=0}^{W-1} \mathbf{X}_F(x, y) e^{j 2\pi \left( x\frac{h}{H} + y\frac{w}{W} \right)}.
\end{equation}

For an intermediate feature map $\feature\in\mathbb{R}^{C\times H\times W}$, its transformation to the frequency space and the corresponding inverse are defined in a similar manner. Unlike input images, the frequency representation of feature maps is denoted as $\feature_f\in\mathbb{R}^{2C\times \lfloor \frac{H}{2} +1 \rfloor \times H}$, where the real and imaginary parts of the FFT output are concatenated along the channel dimension, each occupying $C$ channels. Due to the conjugate symmetric property of the FFT for real-valued inputs, it is sufficient to retain only half of the frequency components along one spatial axis. As a result, the spatial resolution of $\feature$ is reduced to $\lfloor \frac{H}{2} +1 \rfloor \times H$. This representation not only reduces redundancy but also enables direct convolutional operations in the frequency domain, thereby supporting frequency-aware feature manipulation within the network.

\vspace{-0.15cm}\subsection{Overview}
\begin{figure}[th]
    \centering
    \includegraphics[width=1.0\linewidth]{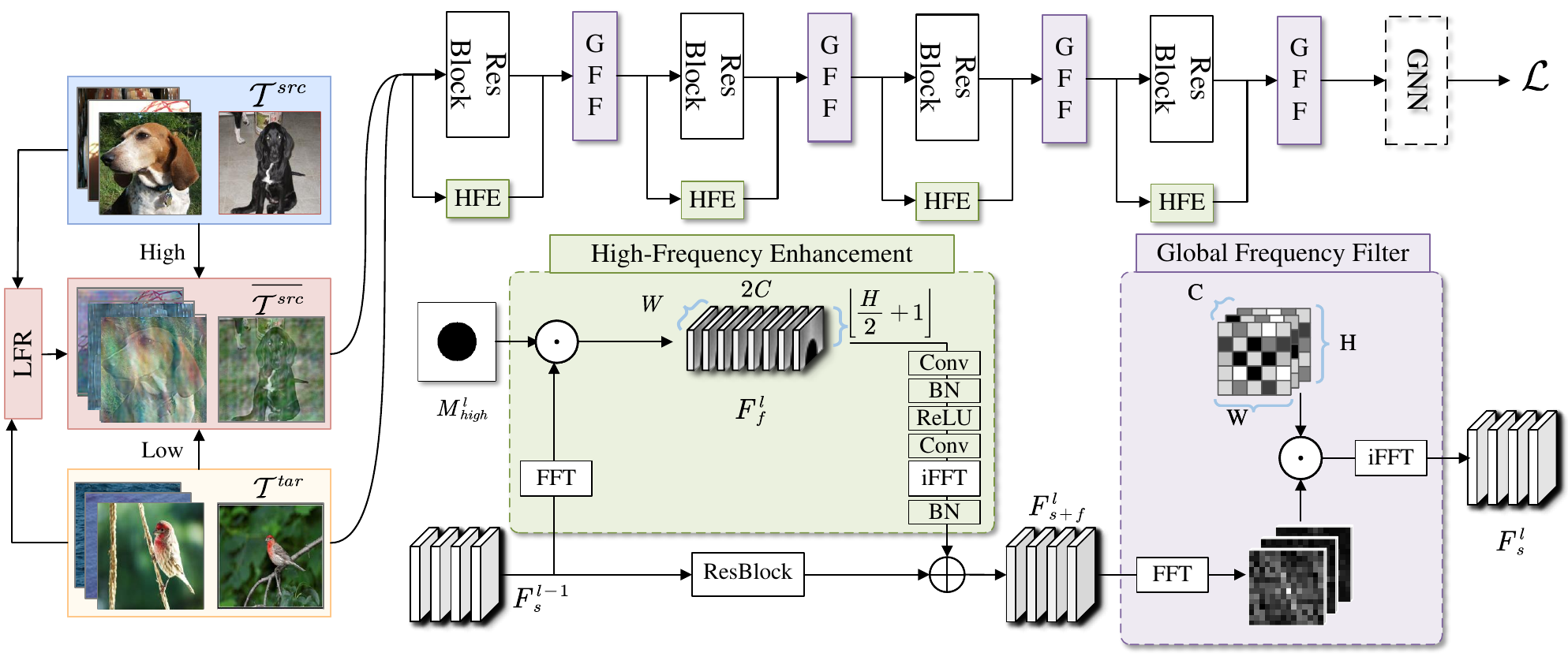}
    \vspace{-1cm}
    \caption{
Overview of the proposed \textbf{FreqGRL} framework, which consists of three key modules: Low-Frequency Replacement (LFR), High-Frequency Enhancement (HFE), and Global Frequency Filter (GFF).}
    \label{fig:framework}
\vspace{-0.5cm} \end{figure}

Figure~\ref{fig:framework} illustrates the overall architecture of the proposed \textbf{FreqGRL} framework, which consists of three core components: \textbf{Low-Frequency Replacement (LFR), High-Frequency Enhancement (HFE), and Global Frequency Filter (GFF)}. \textbf{LFR} mitigates domain shift by replacing the low-frequency components of the source-domain samples with those from the target domain, thereby suppressing the learning of source-specific knowledge. The \textbf{HFE} module enhances the model's ability to focus on domain-generalizable features. It performs a channel-wise 2D FFT on the intermediate feature maps, then selectively retains high-frequency components while masking low-frequency information. The \textbf{GFF} module further refines frequency representations by applying a learnable global filter in the frequency domain, amplifying informative components and suppressing noisy ones. 

Feature extraction is performed through four sequential blocks, each consisting of a residual block, identical in structure to those in ResNet-10. The HFE module is applied within each block to enhance high-frequency information, while GFF is applied after each block to modulate frequency components. The output from the final block is fed into a two-layer graph neural network (GNN) following the structure used in GGCM~\cite{hui2024gradient}, to predict class probabilities for the unlabeled image. Details of the frequency-based modules are provided in the following subsections.

\vspace{-0.15cm}\subsection{Low-frequency replacement}
\begin{figure}[t]
    \centering
    \includegraphics[width=1.0\linewidth]{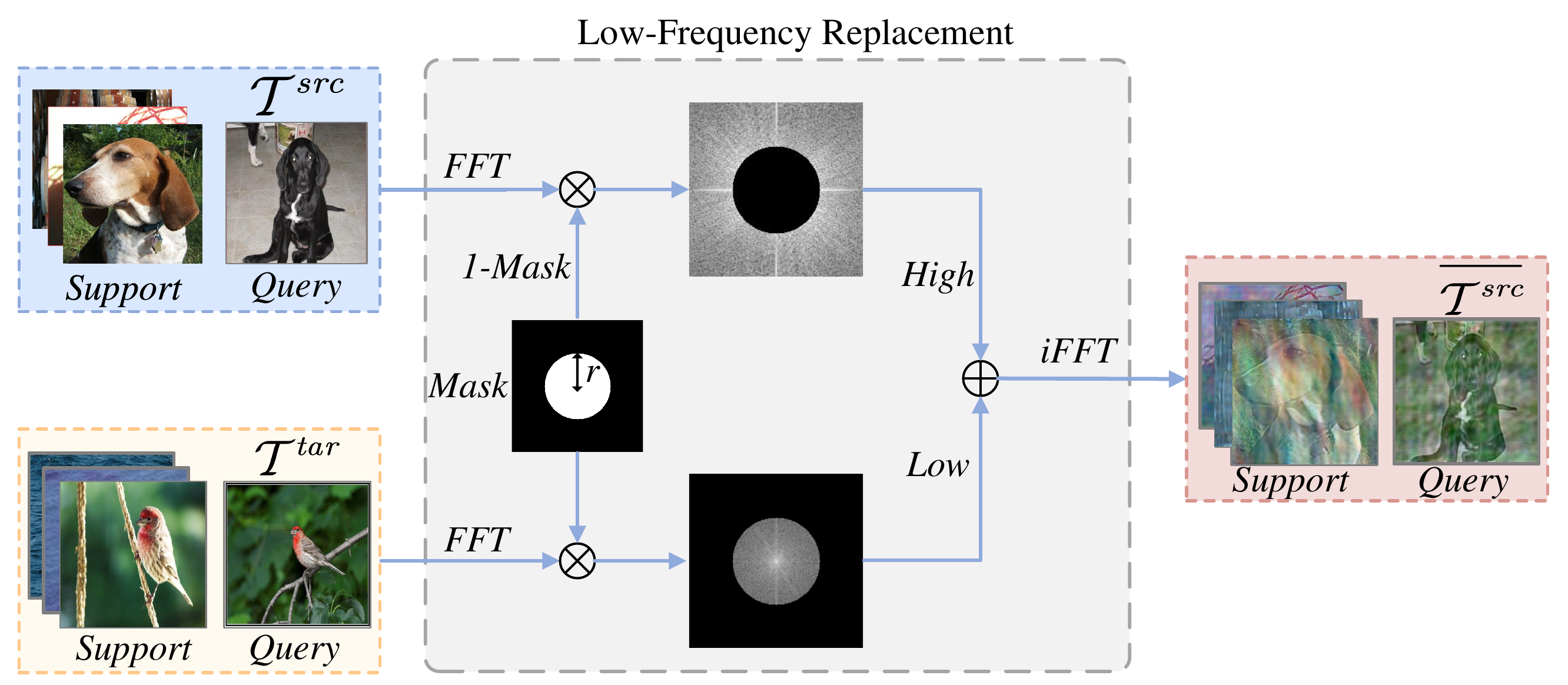}
    \vspace{-1cm}
    \caption{
     \textbf{Illustration of the LFR module}. The source and target tasks are transformed into frequency space and decomposed into the low- and high-frequency components. Then, the source high-frequency and target low-frequency are fused and remapped back to the spatial space to reconstruct the novel source task.
    }
    \label{fig:LFR}
\vspace{-0.5cm} \end{figure}

The LFR module addresses domain shift by replacing the low-frequency components of the source task with those from the target task. As shown in Figure~\ref{fig:LFR}, given a source task $\stask$ and a target task $\ttask$, their images are first transformed into the frequency space via 2D FFT, yielding $\stask_f=FFT(\stask)$ and $\ttask_f=FFT(\ttask)$. The frequency spectrum of the source task is then decomposed into high-frequency and low-frequency components using a binary low-frequency mask $M_{low}\in\{0,1\}^{H\times W}$:
\begin{equation}
    \begin{aligned}
        \stask_{f\_low}&=M_{low}\odot\stask_f \\
        \stask_{f\_high}&=(1-M_{low})\odot\stask_f
    \end{aligned}
\end{equation}
where $\odot$ denotes the broadcasted and element-wise multiplication. The mask $M_{low}\in\{0,1\}^{H\times W}$ is designed to preserve frequency components near the center of the spectrum, defined as:

\[
M_{l(u,v)} =
\left\{
\begin{aligned}
1, &\quad \text{if } \max\left( \left| u - \frac{H}{2} \right|, \left| v - \frac{W}{2} \right| \right) \le r \\
0, &\quad \text{otherwise}
\end{aligned}
\right.
\]

This formulation ensures that only low-frequency components within a low-frequency radius $r$ from the center are retained, while others are suppressed. The radius $r$ is determined by $r=\gamma \cdot min(H, W)$, where $\gamma$ is randomly sampled from a uniform distribution $U(0,0.2)$. Using the same process, the low-frequency and high-frequency components of the target task are obtained. Prior studies~\cite{chen2023learnable,guo2023aloft} have shown that the low-frequency components often carry domain-specific knowledge. To mitigate the learning of source-specific biases, we replace the low-frequency components of the source task with those from the target task, while retaining the original high-frequency components of the source task. The modified frequency representation is computed as: $\overline{\stask_f}=\ttask_{f\_low}+\stask_{f\_high}$.

This modified spectrum is then transformed back into spatial space using the inverse FFT to form the augmented source task: $\overline{\stask}=iFFT(\overline{\stask_f})$.   

\vspace{-0.15cm}\subsection{High-frequency enhancement}
The HFE module is designed to strengthen the model’s focus on high-frequency, domain-generalizable features that are crucial for cross-domain adaptation. Given the intermediate feature maps $\feature^{l-1}_s\in \mathbb{R}^{C\times H \times W}$ from the $l-1$-th block in the spatial space, HFE first projects the feature maps into the frequency space using a 2D FFT. To suppress the influence of low-frequency, domain-specific information, a binary high-frequency mask $M^l_{high}\in\{0,1\}^{C\times H \times W}$ is applied to retain only the high-frequency components, which are typically more transferable across domains. The frequency-masked representation is obtained as:
\begin{equation}
    \feature^{l}_{f}=FFT(\feature^{l-1}_s)\odot M^l_{high},
\end{equation}
where $M^l_{\text{high}}$ is a binary high-pass mask constructed by setting the central region (within a radius of $0.5 \cdot \min(H, W)$) to 0 and the remaining elements to 1, thereby preserving only high-frequency components. To further refine the retained high-frequency signals, we apply a lightweight convolutional network $C(\feature^{l}_{f};\theta^{l})$, where $\theta^l$ denotes the learnable parameters. The architecture of $C(\cdot)$ consists of two convolutional layers: the first uses a $3 \times 3$ kernel followed by Batch Normalization (BN) and ReLU activation, and the second uses a $1 \times 1$ kernel to adjust channel interactions. 

After processing in the frequency domain, the output is transformed back to the spatial domain via inverse FFT (iFFT), followed by an additional BN layer to stabilize the distribution. The resulting high-frequency feature map is then added to the spatial output of the $l$-th residual block to emphasize high-frequency information:
\begin{equation}
    \feature^l_{s+f} = \feature^l_s + BN(iFFT(C(\feature^{l}_{f}; \theta^{l}))).
\end{equation}
This enhanced feature map is then passed to the Global Frequency Filter (GFF) module for further modulation in the frequency domain.

\vspace{-0.15cm}\subsection{Global frequency filter}
The GFF module is designed to further refine feature representations by adaptively modulating frequency components in an end-to-end learnable manner. While the LFR and HFE modules respectively suppress source-specific low-frequency bias and enhance domain-generalizable high-frequency information, GFF acts as a global frequency-aware gating mechanism. It selectively amplifies informative frequencies while suppresses noisy or irrelevant ones, particularly important under limited target-domain supervision. 

Given a feature map $\feature^l_{s+f}\in\mathbb{R}^{C\times H \times W}$, already enhanced by the HFF module, GFF first transforms it into the frequency domain. It then applies a learnable global frequency filter $\mathbf{W}^l\in \mathbb{R}^{C\times H \times W}$, specific to the $l$-th layer. This filter consists of trainable weights that are applied directly to the frequency spectrum via element-wise multiplication: $\feature^l_{filt} = FFT(\feature^l_{s+f})\odot\mathbf{W}^l$.
The filtered frequency features are then projected back to the spatial space for subsequent processing: $\feature^l_s=iFFT(\feature^l_{filt})$.

\vspace{-0.15cm}\subsection{Training and inference}
The model is trained episodically by minimizing few-shot losses over both the pseudo source task $\overline{\mathcal{T}^{src}}$ and the target task $\ttask$. To ensure stable training, the loss on the original source task $\stask$ is also included. The total loss function is defined as: $\mathcal{L}=\overline{\lfsl^{src}}+\lfsl^{tar}+\lfsl^{src}$, where $\overline{\lfsl^{src}}$, $\lfsl^{src}$, and $\lfsl^{tar}$ denote the few-shot losses for the pseudo source, original source, and target tasks, respectively. During inference, each episodic task sampled from the unseen target domain is processed by the trained network using original input images. The final class prediction for each query sample is determined by selecting the class with the highest predicted probability.
\section{Experiment}
In this section, we evaluate the proposed FreqGRL framework and its components on four CD-FSL benchmark datasets. We first detail the experimental setup, including datasets, implementation specifics, evaluation protocols, and baseline comparisons. Next, we benchmark FreqGRL against state-of-the-art methods to demonstrate its effectiveness. Finally, we conduct extensive ablation studies to analyze the individual contributions of analyze the individual contributions of each component within our framework.

\subsection{Experimental settings}
\NB{Datasets.}
We utilize 5 datasets: miniImageNet~\cite{russakovsky2015imagenet}, CUB~\cite{branson2010visual}, Cars~\cite{krause20133d}, Places~\cite{zhou2017places}, and Plantae~\cite{van2018inaturalist}. Among them, the miniImageNet-a widely used subset of ImageNet—is used as the source domain and serves as a standard benchmark dataset for meta-learning and FSL. In line with prior works~\cite{fu2021meta,zhuo2022tgdm,fu2022me}, the remaining four datasets are treated as distinct target domains for cross-domain evaluation. 

Following standard protocols~\cite{fu2021meta,zhuo2022tgdm,fu2022me}, we adopt the conventional dataset splits and introduce only a limited number of labeled samples from each target domain during training. The specific data configurations are summarized in Table~\ref{tab:dataset}. As shown, the amount of labeled target-domain data is significantly smaller than that of the source dataset, reflecting the practical challenges inherent to cross-domain few-shot learning scenarios.

\begin{table}[t]
\tiny
\centering
\caption{Statistics of benchmark datasets used in our experiments.}
\begin{tabular}{cc|ccc}
\hline
\multicolumn{2}{c|}{Datasets}                               & Training classes & Images per class & Training images \\ \hline
\multicolumn{1}{c|}{Source}                  & miniImageNet & 64               & 600              & 38400           \\ \hline
\multicolumn{1}{c|}{\multirow{4}{*}{Target}} & CUB          & 100              & 5                & 500             \\
\multicolumn{1}{c|}{}                        & Cars         & 98               & 5                & 490             \\
\multicolumn{1}{c|}{}                        & Places       & 183              & 5                & 915             \\
\multicolumn{1}{c|}{}                        & Plantae      & 100              & 5                & 500             \\ \hline
\end{tabular}
\label{tab:dataset}
\vspace{-0.5cm} \end{table}

\NB{Implementation details.}
To ensure fair comparisons with prior methods, we adopt ResNet-10 as the feature extractor, following established practice. The backbone comprises four residual blocks with output channel sizes of {64, 128, 256, 512}, and is pretrained on the miniImageNet dataset. For classification, we employ GNN across all experiments. Training is conducted on 5-way 1-shot and 5-way 5-shot tasks, with each task including 16 randomly sampled query images per class. All input images are resized to 224$\times$224, and standard data augmentation techniques—random horizontal flipping and color jittering—are applied. The model is trained for 400 epochs, each consisting of 100 episodes. We use the Adam optimizer with a fixed learning rate of 0.001. All experiments are implemented in Python 3.9 and PyTorch 1.13.1, and executed on a single NVIDIA GeForce RTX 3090 GPU with CUDA 11.1. For fair ablation studies, we also train a baseline model that uses the same ResNet-10 backbone and GNN classifier as our full model, but excludes all proposed modules. Both the baseline and our full model are trained under identical settings.

\NB{Evaluation.}
For evaluation, we randomly sample 1,000 episodic tasks from the novel target datasets and report the average classification accuracy along with the 95\% confidence interval. To ensure statistical robustness, all results are averaged over five independent runs.

\subsection{Comparison with state-of-the-art methods}
We conduct a comprehensive comparison between our proposed method and existing CD-FSL approaches. The compared methods include target-free methods such as LPR~\cite{sun2021explanation}, FWT~\cite{tseng2020cross}, ATA~\cite{wang2021cross}, ProD~\cite{ma2023prod}, WaveSAN~\cite{fu2022wave}, AFA~\cite{hu2022adversarial}, StyleAdv~\cite{fu2023styleadv}, and FLoR~\cite{zou2024flatten}, as well as methods that utilize limited labeled target data, including m-LPR, m-AFA, m-FWT, m-WaveSAN, m-FLoR, FDMixup~\cite{fu2021meta}, ME-D2N~\cite{fu2022me}, TGDM~\cite{zhuo2022tgdm}, and CDCNet~\cite{li2025cdcnet}. The prefix "m-" indicates the modified version of target-free methods, adapted to incorporate auxiliary labeled target-domain samples. All methods use ResNet-10 to ensure fair comparisons.

\begin{table}[t]
\tiny
\centering
\caption{Comparison of 5-way 1-shot classification accuracy (\%) on four target datasets. “TF” refers to target-free methods, while “TA” denotes target-accessible methods that utilize limited number of labeled target samples from the target domain during training. The best performance is highlighted in \textbf{bolded} font.}
\label{tab:1shot}
\begin{tabular}{ccccccc}
\hline
5-way & 1-shot & CUB & Cars & Places & Plantae & Avg \\ \hline
\multirow{8}{*}{TF} 
& LPR        & 48.29 $\pm$ 0.51 & 32.78 $\pm$ 0.39 & 54.83 $\pm$ 0.56 & 37.49 $\pm$ 0.43 & 43.35 \\
& ATA        & 45.00 $\pm$ 0.50 & 33.61 $\pm$ 0.40 & 53.57 $\pm$ 0.50 & 34.42 $\pm$ 0.40 & 41.65 \\
& ProD       & 53.97 $\pm$ 0.71 & 38.02 $\pm$ 0.63 & 53.92 $\pm$ 0.72 & 42.86 $\pm$ 0.59 & 47.19 \\
& AFA        & 41.02 $\pm$ 0.40 & 33.52 $\pm$ 0.40 & 54.66 $\pm$ 0.50 & 37.60 $\pm$ 0.40 & 41.70 \\
& FWT        & 47.47 $\pm$ 0.75 & 31.61 $\pm$ 0.53 & 55.77 $\pm$ 0.79 & 35.95 $\pm$ 0.58 & 42.70 \\
& StyleAdv   & 48.49 $\pm$ 0.72 & 34.64 $\pm$ 0.57 & 58.58 $\pm$ 0.83 & 41.13 $\pm$ 0.67 & 45.71 \\
& WaveSAN    & 50.25 $\pm$ 0.74 & 33.55 $\pm$ 0.61 & 57.75 $\pm$ 0.82 & 40.71 $\pm$ 0.66 & 45.57 \\
& FLoR       & 50.38 $\pm$ 0.73 & 36.96 $\pm$ 0.58 & 52.17 $\pm$ 0.70 & 39.01 $\pm$ 0.59 & 44.63 \\
\hline
\multirow{12}{*}{TA} 
& m-LPR      & 59.23 $\pm$ 0.58 & 46.88 $\pm$ 0.53 & 57.92 $\pm$ 0.58 & 49.11 $\pm$ 0.54 & 53.29 \\
& m-AFA      & 54.22 $\pm$ 0.79 & 45.70 $\pm$ 0.69 & 58.42 $\pm$ 0.83 & 45.67 $\pm$ 0.75 & 51.00 \\
& m-FWT      & 61.16 $\pm$ 0.81 & 49.01 $\pm$ 0.76 & 57.89 $\pm$ 0.82 & 50.49 $\pm$ 0.81 & 54.64 \\
& m-StyleAdv & 61.26 $\pm$ 0.83 & 49.67 $\pm$ 0.77 & 61.97 $\pm$ 0.85 & 50.43 $\pm$ 0.79 & 55.83 \\
& m-WaveSAN  & 63.59 $\pm$ 0.85 & 50.06 $\pm$ 0.76 & 59.89 $\pm$ 0.86 & 51.99 $\pm$ 0.81 & 56.38 \\
& m-FLoR     & 55.20 $\pm$ 0.84 & 40.16 $\pm$ 0.71 & 56.17 $\pm$ 0.79 & 42.90 $\pm$ 0.78 & 48.61 \\
& FDMixup    & 63.24 $\pm$ 0.82 & 51.31 $\pm$ 0.83 & 58.22 $\pm$ 0.82 & 51.03 $\pm$ 0.81 & 55.95 \\
& ME-D2N     & 65.05 $\pm$ 0.83 & 49.53 $\pm$ 0.79 & 60.36 $\pm$ 0.86 & 52.89 $\pm$ 0.83 & 56.96 \\
& TGDM       & 64.80 $\pm$ 0.26 & 50.70 $\pm$ 0.24 & 61.88 $\pm$ 0.26 & 52.39 $\pm$ 0.25 & 57.44 \\
& CDCNet     & 66.55 $\pm$ 0.83 & 52.85 $\pm$ 0.84 & 62.73 $\pm$ 0.83 & 52.67 $\pm$ 0.76 & 58.70 \\
& GGCM       & 68.93 $\pm$ 0.83 & 56.43 $\pm$ 0.87 & 63.63 $\pm$ 0.87 & 54.34 $\pm$ 0.87 & 60.83 \\
& MetaCM     & 67.68 $\pm$ 0.82 & 55.43 $\pm$ 0.87 & 63.52 $\pm$ 0.86 & 55.11 $\pm$ 0.86 & 60.44 \\
\hline
\multirow{2}{*}{Ours} 
& Baseline   & 57.99 $\pm$ 0.79 & 42.51 $\pm$ 0.70 & 57.17 $\pm$ 0.81 & 47.88 $\pm$ 0.75 & 51.39 \\
& FreqGRL    & \textbf{70.29 $\pm$ 0.80} & \textbf{56.77 $\pm$ 0.81} & \textbf{63.63 $\pm$ 0.86} & \textbf{56.94 $\pm$ 0.86} & \textbf{61.91} \\
\hline
\end{tabular}
\vspace{-0.5cm} \end{table}

\begin{table}[t]
\tiny
\centering
\caption{Comparison of 5-way 5-shot classification accuracy (\%) on four target datasets. “TF” refers to target-free methods, while “TA” denotes target-accessible methods that utilize limited number of labeled target samples from the target domain during training. The best performance is highlighted in \textbf{bolded} font.}
\label{tab:5shot}
\begin{tabular}{ccccccc}
\hline
5-way                 & 1-shot     & CUB                   & Cars                  & Places                & Plantae               & Avg            \\ \hline
\multirow{8}{*}{TF}   & LPR        & 64.44 ± 0.48          & 46.20 ± 0.46          & 74.45 ± 0.47          & 54.46 ± 0.46          & 59.89          \\
                      & ATA        & 66.22 ± 0.50          & 49.14 ± 0.40          & 75.48 ± 0.40          & 52.69 ± 0.40          & 60.88          \\
                      & ProD       & 79.19 ± 0.59          & 59.49 ± 0.68          & 75.00 ± 0.72          & 65.82 ± 0.65          & 69.88          \\
                      & AFA        & 59.46 ± 0.40          & 46.13 ± 0.40          & 68.87 ± 0.40          & 52.43 ± 0.40          & 56.72          \\
                      & FWT        & 66.98 ± 0.68          & 44.90 ± 0.64          & 73.94 ± 0.67          & 53.85 ± 0.62          & 59.92          \\
                      & StyleAdv   & 68.72 ± 0.67          & 50.13 ± 0.68          & 77.73 ± 0.62          & 61.52 ± 0.68          & 64.53          \\
                      & WaveSAN    & 70.31 ± 0.67          & 46.11 ± 0.66          & 76.88 ± 0.63          & 57.72 ± 0.64          & 62.76          \\
                      & FLoR       & 73.40 ± 0.69          & 55.58 ± 0.67          & 76.00 ± 0.71          & 59.79 ± 0.72          & 66.19          \\ \hline
\multirow{12}{*}{TA}  & m-LPR      & 77.07 ± 0.44          & 64.38 ± 0.48          & 77.73 ± 0.45          & 67.90 ± 0.47          & 71.77          \\
                      & m-AFA      & 74.55 ± 0.66          & 63.80 ± 0.70          & 79.19 ± 0.60          & 64.25 ± 0.66          & 70.45          \\
                      & m-FWT      & 79.14 ± 0.62          & 65.42 ± 0.70          & 78.59 ± 0.60          & 68.26 ± 0.68          & 72.85          \\
                      & m-StyleAdv & 77.40 ± 0.69          & 64.73 ± 0.72          & 79.86 ± 0.61          & 68.85 ± 0.67          & 72.71          \\
                      & m-WaveSAN  & 82.29 ± 0.58          & 66.93 ± 0.71          & 80.01 ± 0.60          & 71.27 ± 0.70          & 75.13          \\
                      & m-FLoR     & 73.40 ± 0.69          & 55.58 ± 0.67          & 76.00 ± 0.71          & 59.79 ± 0.72          & 66.19          \\
                      & FDMixup    & 79.46 ± 0.63          & 66.52 ± 0.70          & 78.92 ± 0.63          & 69.22 ± 0.65          & 73.53          \\
                      & ME-D2N     & 83.17 ± 0.56          & 69.17 ± 0.68          & 80.45 ± 0.62          & 72.87 ± 0.67          & 76.42          \\
                      & TGDM       & 84.21 ± 0.18          & 70.99 ± 0.21          & 81.62 ± 0.19          & 71.78 ± 0.22          & 77.15          \\
                      & CDCNet     & 84.67 ± 0.57          & 70.08 ± 0.69          & 80.77 ± 0.63          & 71.65 ± 0.70          & 76.79          \\
                      & GGCM       & 85.08 ± 0.54          & 73.01 ± 0.71          & 82.13 ± 0.59          & 74.53 ± 0.67          & 78.69          \\
                      & MetaCM     & 85.59 ± 0.57          & 73.87 ± 0.68          & 81.77 ± 0.61          & 73.62 ± 0.68          & 78.71          \\ \hline
\multirow{2}{*}{Ours} & Baseline   & 81.06 ± 0.58          & 61.16 ± 0.71          & 76.30 ± 0.67          & 67.15 ± 0.68          & 71.42          \\
                      & FreqGRL    & \textbf{88.02 ± 0.52} & \textbf{76.56 ± 0.66} & \textbf{82.24 ± 0.60} & \textbf{78.00 ± 0.66} & \textbf{81.21} \\ \hline
\end{tabular}
\vspace{-0.5cm}
\end{table}

\NB{Comparison results.}
Table~\ref{tab:1shot} and Table~\ref{tab:5shot} present the performance comparisons between our method and existing CD-FSL approaches under the 5-way 1-shot and 5-way 5-shot settings across four benchmark datasets. Our FreqGRL consistently outperforms all baselines, achieving the highest accuracy in every domain and under both evaluation settings. In particular, it improves upon the previous state-of-the-art (SOTA) methods by an average margin of 1.08\% in the 1-shot setting and 2.50\% in the 5-shot setting. Although the improvement in the 1-shot case is relatively modest, this is expected, as FreqGRL benefits more from a larger support set where high-level semantic cues are more stable, enabling more effective exploitation of high-frequency information for cross-domain generalization.

Notably, our method achieves 81.21\% average accuracy in the 5-shot setting—the first CD-FSL method to surpass the 80\% threshold. Similarly, in the 1-shot scenario, it achieves 70.29\% accuracy on the CUB dataset, also setting a new benchmark in this domain. These results highlight the effectiveness of our approach in learning domain-invariant representations and its robustness across heterogeneous target domains, especially under the constraint of extremely limited target-domain data.

Furthermore, we observe that all target-accessible methods outperform their target-free counterparts, reaffirming the value of utilizing even a small number of labeled target samples during training. This supports our design choice of operating in the target-accessible setting to enhance adaptation capability in real-world CD-FSL applications.

\subsection{Ablation study.}
To further verify the effectiveness and design rationale of the proposed FreqGRL framework, we conduct a series of ablation experiments focused on its three core modules: LFR, HFE, and GFF. All ablation studies are performed on the CUB dataset under the 5-way 1-shot setting, using ResNet-10 as the backbone. This setting allows us to isolate and analyze the contribution of each component in a controlled yet challenging CD-FSL scenario.

\NB{Module removal experiments.}
To assess the individual contribution of each component in the proposed framework, we conduct a series of module removal experiments. The results are summarized in Table~\ref{tab:ablate}. Compared to the baseline, Experiments 1, 2, and 3—each incorporating a single module (LFR, HFE, or GFF)—demonstrate noticeable accuracy improvements, confirming that each module independently enhances model performance. Moreover, combining any two modules (Experiments 4, 5, and 6) leads to further gains, indicating their complementary nature. The best performance is observed when all modules are integrated, validating the synergistic effect of the full FreqGRL framework.

\begin{table}[t]
\tiny
\centering
\label{tab:ablate}
\caption{\textbf{Ablation study of module removal experiments}. This table shows the accuracy (mean ± standard deviation) of the model under different module combinations, with the best performance in \textbf{bolded} font.}
\begin{tabular*}{\textwidth}{@{\extracolsep{\fill}} ccccc }
\hline
Experiment & LFR & HFE & GFF & Accuracy         \\ \hline
Baseline   &    &    &    & 57.99 $\pm$ 0.79 \\ \hline
1          & $\checkmark$  &    &    & 64.06 $\pm$ 0.82 \\
2          &    & $\checkmark$  &    & 62.27 $\pm$ 0.80 \\
3          &    &    & $\checkmark$  & 63.18 $\pm$ 0.81 \\
4          & $\checkmark$  & $\checkmark$  &    & 66.27 $\pm$ 0.82 \\
5          & $\checkmark$  &    & $\checkmark$  & 67.37 $\pm$ 0.80 \\
6          &    & $\checkmark$  & $\checkmark$  & 66.48 $\pm$ 0.80 \\ \hline
FreqGRL (Ours)       & $\checkmark$  & $\checkmark$  & $\checkmark$  & \textbf{70.29 $\pm$ 0.80} \\ \hline
\end{tabular*}
\vspace{-0.5cm} \end{table}

\begin{table}[t]\tiny
\centering
\caption{Impact of varying frequency replacement ranges in LFR controlled by $\gamma$.}
\begin{tabular}{c c c}
\hline
Type         & Experiment      & Accuracy         \\ \hline
\multirow{2}{*}{Fixed}  
             & $\gamma=0.1$ &  61.43 $\pm$ 0.81 \\
             & $\gamma=0.2$ &  63.29 $\pm$ 0.82 \\\hline
\multirow{2}{*}{Random} 
             & HFR        & 59.05 $\pm$ 0.80 \\
             & $\gamma\sim U(0,1)$         & 63.66 $\pm$ 0.80 \\ \hline
Ours         & $\gamma\sim U(0,0.2)$         & \textbf{64.06 $\pm$ 0.82} \\ \hline
\end{tabular}
\label{tab:gamma}
\vspace{-0.5cm} \end{table}

\NB{Effect of the replacement region of the LFR.}
To explore the impact of the low-frequency replacement range in the LFR module, we conduct experiments by varying the hyperparameter $\gamma$, which controls the size of the replaced frequency region. We evaluate both fixed and random strategies for setting $\gamma$. In the \textbf{fixed strategy}, $\gamma$ is set to constant values of 0.1 and 0.2. In the \textbf{random strategy}, $\gamma$ is sampled from a uniform distribution, i.e., $\gamma \sim U(0,1)$. To further assess the importance of replacing low-frequency components, we introduce a variant termed \textbf{HFR}, which replaces the \textbf{high-frequency components} of the source task instead. For fair comparison, HFR adopts the same sampling range for $\gamma$ as LFR, i.e., $\ gamma\sim U(0,0.2)$.

The results are summarized in Table~\ref{tab:gamma}. We observe that random sampling strategies outperform fixed ones, indicating that increasing the diversity of replaced frequency components enables the model to better generalize across a wider range of target distributions. Among them, our proposed setting $\gamma \sim U(0,0.2)$ achieves the best performance 64.06\%, suggesting that controlling the replacement range is crucial. A moderate replacement region not only suppresses the low-frequency bias inherent in the source domain but also preserves sufficient high-frequency cues essential for effective representation learning. In contrast, excessively large replacement regions, such as $\gamma \sim U(0,1)$, tend to impair important structural information, leading to performance degradation. Moreover, the HFR variant, which replaces high-frequency components instead of low-frequency ones, results in a significant drop to 59.05\%, further validating our design choice of focusing on low-frequency replacement. These findings demonstrate the effectiveness of the LFR module and highlight the importance of carefully designing the replacement region to strike a balance between diversity and information preservation.

\NB{Ablate study of HFE.}
\begin{figure}[t]
    \centering
    \includegraphics[width=0.65\textwidth]{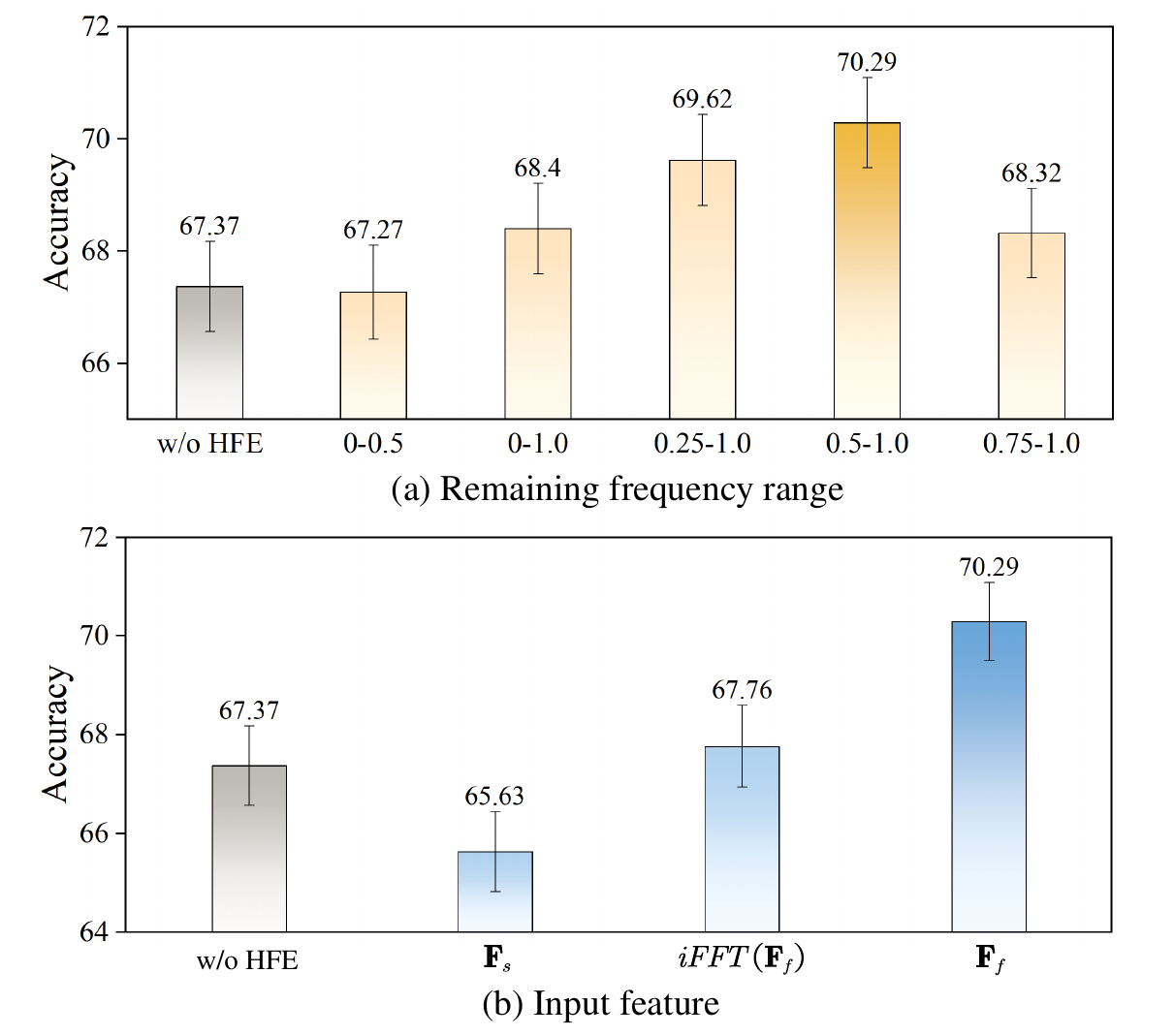}
    \vspace{-0.5cm}
    \caption{\textbf{Ablation study of the HFE module.} (a) Accuracy under different frequency ranges. (b) Accuracy with different input types.}
    \label{fig:hfe_ablation}
\vspace{-0.5cm} \end{figure}
To assess the sensitivity of HFE to the preserved frequency range, we conduct an ablation study by varying the retained high-frequency bands. As illustrated in Figure~\ref{fig:hfe_ablation}(a),  incorporating HFE generally improves performance over the baseline without HFE (67.37\%, shown in gray). However, when only the low-frequency band (0–0.5) is preserved, performance drops slightly to 67.27\%, underscoring the necessity of enhancing high-frequency components. Notably, retaining the 0.5–1.0 frequency band yields the highest accuracy (70.29\%), indicating that mid-to-high frequency are most effective for cross-domain adaptation. In contrast, overly broad ranges like 0–1.0 or overly narrow ones such as 0.75–1.0 lead to diminished performance, likely due to the inclusion of source-specific low-frequency bias or the exclusion of critical mid-frequency cues. These foundings demonstrate that preserving the 0.5–1.0 range provides an optimal balance, and this setting is adopted throughout other experiments. 

To further validate the design of the HFE module, we investigate the impact of different input feature types on the convolutional operations within HFE. Specifically, we compare three inpit variants: the original spatial-domain features without masking ($\feature_s$), the spatial-domain features obtained by inverse FFT of high-frequency-masked frequency features ($iFFT(\feature_f)$), and the high-frequency-masked features directly processed in the frequency domain ($\feature_f$). As shown in Figure~\ref{fig:hfe_ablation}(b), using $\feature_f$—i.e., performing convolution directly in the frequency domain—achieves the highest accuracy of 70.29\%, clearly outperforming the other alternatives. This demonstrates that frequency-domain learning, especially on explicitly enhanced high-frequency signals, effectively captures domain-invariant patterns essential for cross-domain adaptation.  Interestingly, using $iFFT(\feature_f)$ also leads to imporved performance (67.76\%) compared to the baseline wuthout HFE (67.37\%), indicating that applying high-frequency masking is beneficial even when processing is performed in the spatial domain. In contrast, directly using the unmasked spatial features ($\feature_s$) results in the lowest accuracy (65.63\%), reinforcing our motivation that raw spatial features tend to contain domain-specific low-frequency biases detrimental to generalization. In terms of computational cost, although frequency-domain convolution doubles the number of channels due to complex-valued representations, the overall cost remains comparable to that of spatial convolution. This is because the conjugate symmetry of FFT reduces the number of unique spatial positions by half. Therefore, our frequency-domain design not only yields better generalization but also maintains computationally efficiency.

\subsection{Further analysis.}
\begin{table}[t]\scriptsize
\centering
\label{tab:dg}
\caption{Maximum Mean Discrepancy (MMD) between source and target features on four target domains. Lower values indicate smaller domain gaps.}
\begin{tabular}{c|cccc}
\hline
 Method        & CUB      & Cars     & Places   & Plantae  \\ \hline
Baseline & 0.1898 & 0.3242 & 0.1710 & 0.3138 \\
Baseline+LFR     & 0.1451 & 0.2824  & 0.1252 & 0.1322 \\ \hline
\end{tabular}
\vspace{-0.5cm} \end{table}

\begin{figure}[th]
    \centering
    \includegraphics[width=0.7\textwidth]{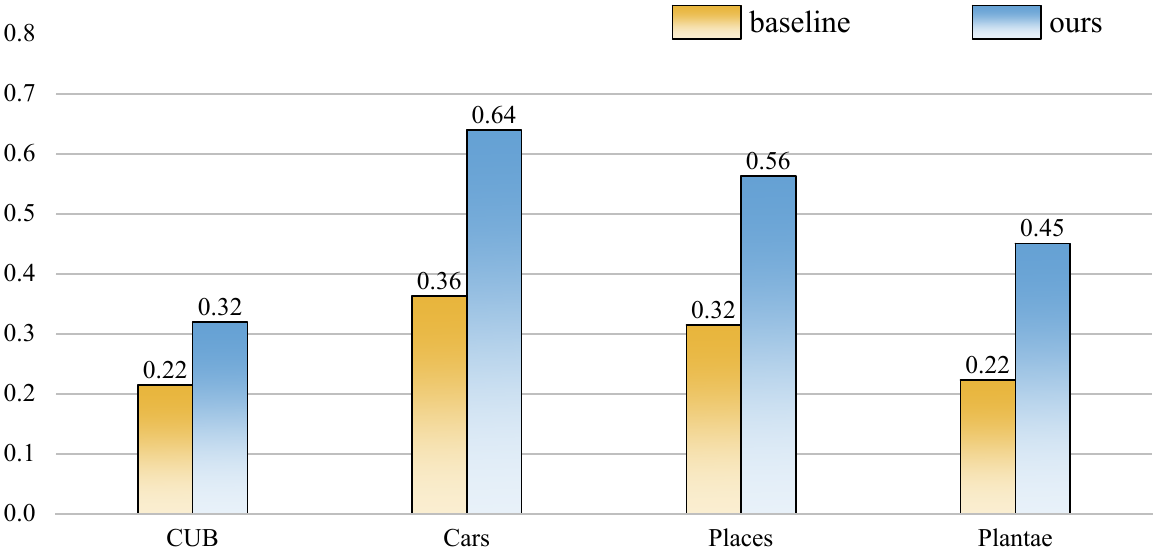} 
    \vspace{-0.5cm}
    \caption{
    Comparison of target high-frequency accuracy ratio across four datasets. 
    }
    \label{fig:target_freq_accuracy}
\vspace{-0.5cm} \end{figure}

\NB{Domain gaps.}  To quantify the distributional shift between source and target domains, we employ Maximum Mean Discrepancy (MMD), where smaller values indicate narrower domain gaps. To evaluate the effectiveness of the proposed LFR module, we compute the MMD between source and target features extracted from the final residual block of the backbone network, both before and after integrating LFR into the baseline model. As shown in Table~\ref{tab:dg}, the incorporation of LFR consistently reduces the domain distance across all four target domains. This suggests that LFR effectively suppresses source-specific patterns and constructs source tasks that are more aligned with the target domain, thereby facilitating better generalization.

\NB{Comparison of the target high-frequency accuracy ratio.}
To further validate the effectiveness of the HFE module in enhancing domain-generalizable frequency features, we compare the target high-frequency accuracy ratio on the target domain between our method and the baseline across four target datasets: CUB, Cars, Places, and Plantae. As shown in Figure~\ref{fig:target_freq_accuracy}, our method consistently achieves higher high-frequency accuracy ratios than the baseline across all domains, demonstrating the ability of the proposed HFE module to enhance the discriminative power of transferable high-frequency components, thereby contributing to improved cross-domain generalization.

\begin{figure}[t]
    \centering
    \includegraphics[width=0.7\textwidth]{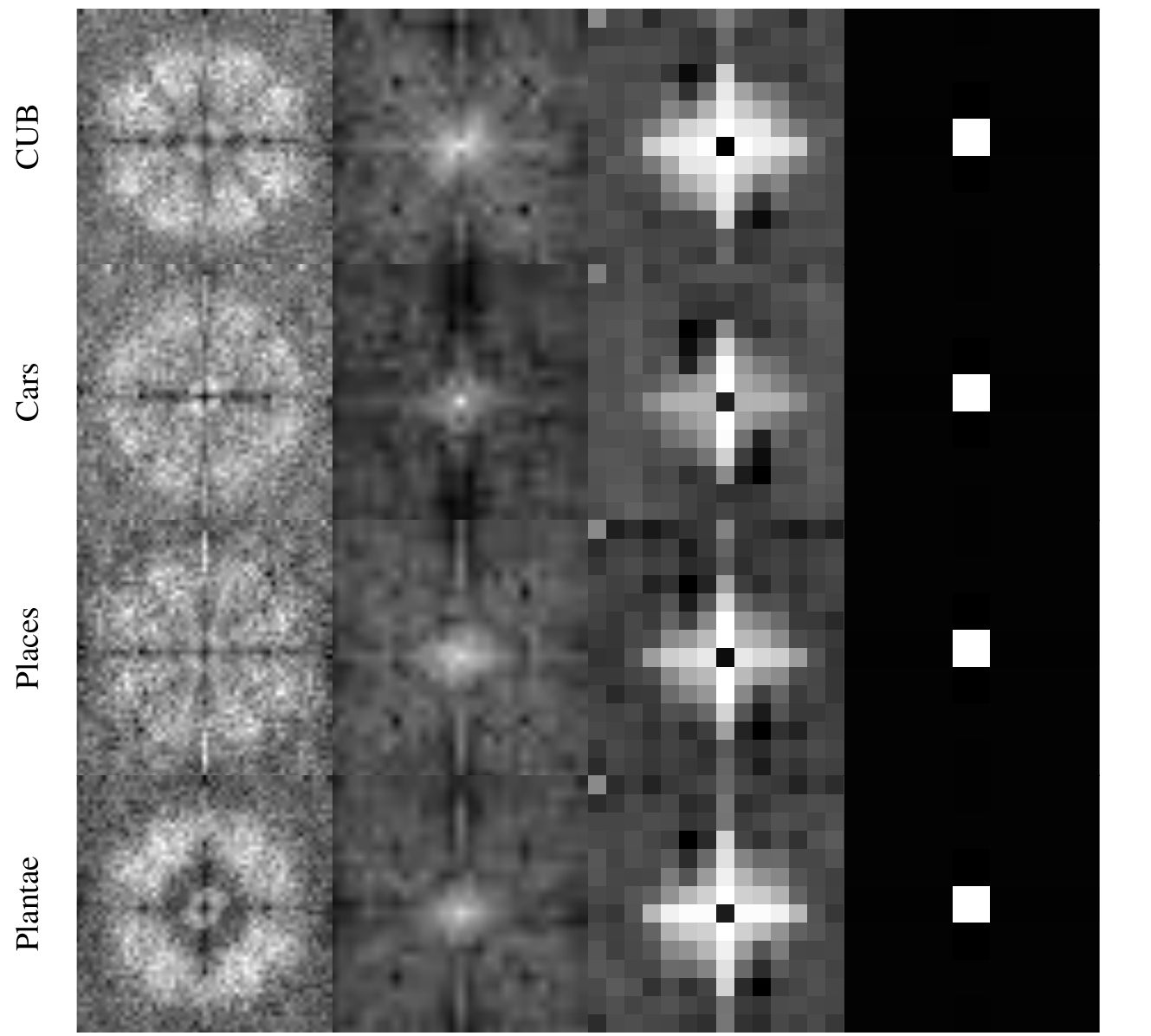}
    \vspace{-0.5cm}
    \caption{
    \textbf{Visualization of frequency weights learned by GFF modules} across four target datasets (CUB, Cars, Places, and Plantae). In each row, we show the learned frequency weight maps from the 1st to 4th GFF block. All maps are obtained by averaging over the channel dimension and min-max normalized to the range [0, 1].}
    \label{fig:gff}
\vspace{-0.5cm} \end{figure}

\NB{Visualization of GFF filters.}
To better understand the behavior of the proposed GFF module, we visualize the learned frequency weight maps (i.e., the magnitude spectrum of FFT-weighted responses) from each residual block across target datasets. As shown in Figure~\ref{fig:gff}, we make the following key observations:  First, Low-frequency components are generally assigned higher weights, while high-frequency components are suppressed across all datasets. This suggests that high-frequency regions tend to be noisier under the few-shot learning setting, where limited samples make it difficult to extract stable and reliable high-frequency cues. GFF adaptively downweights these noisy components, thereby improving generalization. Second, we observe that low-frequency weights exhibit greater variation across datasets compared to high-frequency ones. This indicates that low-frequency regions encode more domain-specific biases. The GFF module learns to selectively suppress such domain-specific low-frequency signals while preserving transferable ones, effectively aligning feature distributions across domains. Finally, in the 4th residual block, the learned frequency weights exhibit an extremely concentrated low-pass pattern, where only the lowest frequency component is modulated while the others remain uniform (i.e., assigned a value of 1). This behavior is consistent with the global average pooling (GAP) operation applied after this block. Since GAP inherently acts as a low-pass filter, only the lowest frequency components contribute to the training signal at this stage, leaving higher frequency components unaffected.

\begin{table}[t]
\tiny
\centering
\caption{Comparison of the computational complexity and accuracy.}
\label{tab:comp}
\begin{tabular}{cccc}
\hline
         & FLOPS (G) & Parameters (M) & Avg. acc (\%) \\ \hline
Baseline & 8.408     & 5.383          & 51.39         \\
FDMixup  & 8.420      & 5.581          & 55.95         \\
ME-D2N   & 16.82     & 5.385          & 56.96         \\
Ours     & 16.70      & 7.696          & 61.91         \\ \hline
\end{tabular}
\vspace{-0.5cm} \end{table}
\NB{Complexity Analysis.}
To assess the computational overhead introduced by our frequency-aware modules, we compare the inference-time complexity of our method with that of the baseline model and two state-of-the-art approaches (FDMixup and ME-D2N) under the 5-way 1-shot setting. All methods adopt the same ResNet-10 backbone and GNN classifier to ensure fair comparison.

As reported in Table~\ref{tab:comp}, our method introduces a moderate increase in parameters (7.696M vs. 5.385M for the baseline), primarily due to the additional HFE and GFF modules. Nonetheless, the total FLOPs (16.70G) remain comparable to ME-D2N (16.82G), while our method achieves a significantly higher average accuracy (61.91\% vs. 56.96\%).  These results confirm that the performance gain afforded by our frequency-guided design comes at only moderate computational cost, demonstrating a favorable trade-off between accuracy and complexity.
\section{Conclusions}
In this paper, we present a novel frequency-space perspective to address the challenges of data imbalance in CD-FSL with limited target-domain supervision. Our analysis reveals that excessive source data can lead to overfitting on low-frequency, domain-specific patterns, while the sparsity of target data impairs the learning of high-frequency, domain-generalizable features. To mitigate these issues, we propose FreqGen, a frequency-aware CD-FSL framework. The LFR module substitutes the low-frequency components of source data with those from the target domain to reduce source bias. The HFE module amplifies high-frequency features in the frequency space to promote domain-invariant knowledge learning. Moreover, the GFF adaptively modulates frequency responses to suppress noise and enhance task-relevant features. Extensive experiments across multiple benchmarks validate the effectiveness of our method and demonstrate its superiority over existing approach.

\section{Acknowledgements}
This work is supported in part by National Science and Technology Major Project under Grants 2023ZD0121300, NSFC under Grants 62088102, 12326608, and 62106192, Natural Science Foundation of Shaanxi Province under Grant 2022JC-41, Fundamental Research Funds for the Central Universities under Grant XTR042021005, and the Public Welfare Research Program of Ningbo City under Grant 2024S063.




\setlength{\bibsep}{0em}
\bibliographystyle{elsarticle-num-names}\scriptsize
\bibliography{ref}






\end{document}